\documentclass[10pt,twocolumn,letterpaper]{article}

\usepackage{iccv}
\usepackage{times}
\usepackage{epsfig}
\usepackage{graphicx}
\usepackage{amsmath}
\usepackage{amssymb}
\usepackage[accsupp]{axessibility} 
\usepackage[nocompress]{cite}

\makeatletter
\DeclareRobustCommand\onedot{\futurelet\@let@token\@onedot}
\def\@onedot{\ifx\@let@token.\else.\null\fi\xspace}

\makeatother

\newcommand{\hide}[1]{}

\usepackage{color}
\usepackage{xcolor,colortbl}
\usepackage{mathtools}
\usepackage{multirow}
\usepackage{bm} 
\usepackage{pifont}
\usepackage{xspace}
\usepackage{booktabs}
\definecolor{purple}{rgb}{0.65,0,0.65}
\definecolor{dark_green}{rgb}{0, 0.5, 0}
\definecolor{blueish}{rgb}{0.0, 0.3, .6}
\definecolor{orange}{rgb}{0.9, 0.5, 0}
\definecolor{tabhighlight}{HTML}{e5e5e5}

\newcommand{\cmark}{\ding{51}}%
\newcommand{\method}{ISSUES\xspace}

\newcommand\blfootnote[1]{%
  \begingroup
  \renewcommand\thefootnote{}\footnote{#1}%
  \addtocounter{footnote}{-1}%
  \endgroup
}

\usepackage[pagebackref=true,breaklinks=true,letterpaper=true,colorlinks,bookmarks=false]{hyperref}

\usepackage[capitalize]{cleveref}
\crefname{section}{Sec.}{Secs.}
\Crefname{section}{Section}{Sections}
\Crefname{table}{Table}{Tables}
\crefname{table}{Tab.}{Tabs.}

\iccvfinalcopy 

\def\iccvPaperID{27} 

\ificcvfinal\fi

\begin{document}

\title{Mapping Memes to Words for Multimodal Hateful Meme Classification}

\author{
\textsuperscript{$*$}Giovanni Burbi\textsuperscript{1} \and
\textsuperscript{$*$}Alberto Baldrati\textsuperscript{1,2} \and Lorenzo Agnolucci\textsuperscript{1} \and  Marco Bertini\textsuperscript{1} \and Alberto Del Bimbo\textsuperscript{1} \and
\textsuperscript{1} University of Florence - Media Integration and Communication Center (MICC) \\  \textsuperscript{2} University of Pisa \\ 
Florence, Italy - Pisa, Italy\\
{\tt\small [name.surname]@unifi.it}
}

\maketitle
\ificcvfinal\thispagestyle{empty}\fi

\begin{abstract}
Multimodal image-text memes are prevalent on the internet, serving as a unique form of communication that combines visual and textual elements to convey humor, ideas, or emotions. However, some memes take a malicious turn, promoting hateful content and perpetuating discrimination. Detecting hateful memes within this multimodal context is a challenging task that requires understanding the intertwined meaning of text and images. In this work, we address this issue by proposing a novel approach named \method for multimodal hateful meme classification. \method leverages a pre-trained CLIP vision-language model and the textual inversion technique to effectively capture the multimodal semantic content of the memes. The experiments show that our method achieves state-of-the-art results on the Hateful Memes Challenge and HarMeme datasets. The code and the pre-trained models are publicly available at 
\href{https://github.com/miccunifi/ISSUES}{\small{\url{https://github.com/miccunifi/ISSUES}}}
\blfootnote{\textsuperscript{$*$}~Equal contribution}

\textbf{Disclaimer: This paper contains hateful content that may be disturbing to some readers.}
\end{abstract}

\begin{figure*}
  \centering
  \includegraphics[width=0.9\textwidth]{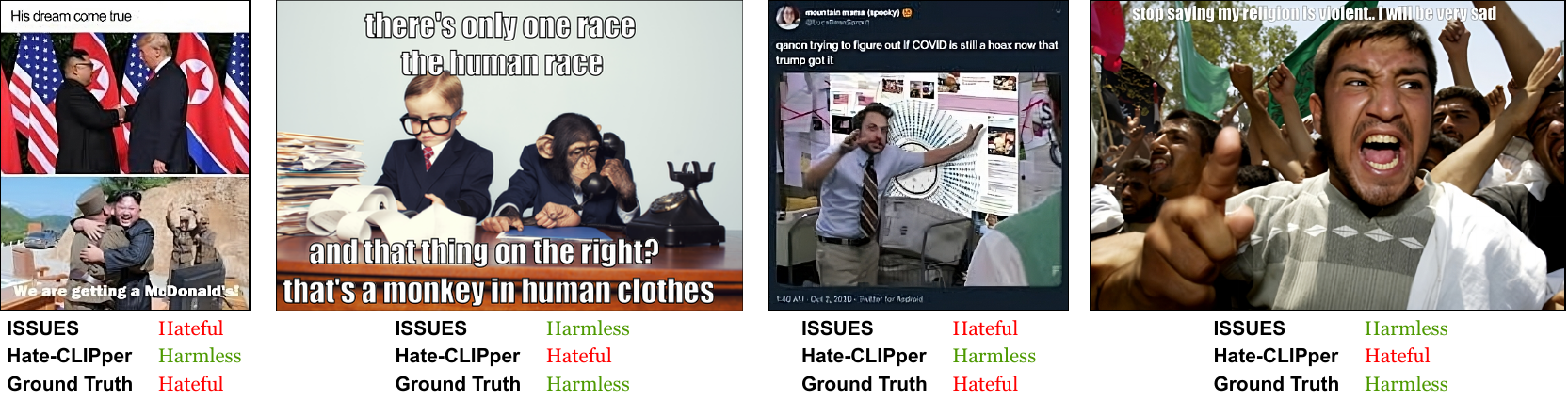}
  \caption{Examples of multimodal image-text memes. Given a meme, we want to classify whether its content conveys hate. The proposed \method approach is more effective at evaluating the hatefulness of the memes than the state-of-the-art method Hate-CLIPper.} 
  \vspace{-2pt}
  \label{fig:teaser}
\end{figure*}

\vspace{-8pt}
\section{Introduction}
Multimodal image-text memes are a unique form of memes that combine visual elements and textual content to convey humor, ideas, or emotions. Unfortunately, some of them are used to perpetuate discrimination against individuals or groups based on their identity \cite{kiela2020hateful}. Classifying hateful memes is particularly challenging, as their true intent subtly emerges only when text and images are integrated. \Cref{fig:teaser} shows some examples of multimodal memes where innocuous images and texts turn hateful when combined together. 

The Hateful Memes Challenge~\cite{kiela2020hateful} (HMC) played a pivotal role in advancing research on automated hateful meme classification. The challenge presented curated memes where visual and textual information were tightly intertwined, making the use of multimodal approaches essential. In particular, the organizers crafted non-hateful "confounder" memes by altering only the image or text in the hateful memes while preserving their overall context. These confounder memes show that a seemingly harmless image or text could turn hateful depending on the contextual cues present in the other modality. 

To tackle the hateful meme classification task we present a novel approach named \method (mappIng memeS to wordS for mUltimodal mEme claSsification) that leverages a pre-trained CLIP vision-language model and the recently introduced textual inversion technique~\cite{gal2022image, baldrati2023zero}. Following the terminology introduced in \cite{gal2022image}, we refer to \textit {textual inversion} as the process of mapping an image into a pseudo-word token residing in the CLIP token embedding space. 
\method introduces a powerful framework based on three key concepts. First, by exploiting the textual inversion technique, we enhance the multimodal capabilities of the model, allowing the creation of a multimodal representation within the textual embedding space. Second, we disentangle image and text features and we adapt both embedding spaces to the specific downstream task. Finally, inspired by \cite{baldrati2022effective}, we design an effective multimodal fusion network.
Experiments show that our approach achieves SotA results on the challenging HMC~\cite{kiela2020hateful} dataset and on the HarMeme~\cite{pramanick2021detecting} dataset.

We summarize our contributions as follows:
\begin{itemize}
\vspace{-6pt}
    \item We introduce \method, a novel approach for multimodal meme classification which leverages textual inversion in conjunction with a frozen pre-trained CLIP vision-language model;
    \vspace{-6pt}
    \item To the best of our knowledge, this is the first work that shows that textual inversion can be effectively used to enrich the textual features in a classification task;
    \vspace{-6pt}
    \item The proposed method achieves SotA results on two datasets: Hateful Memes Challenge and HarMeme.
\end{itemize}

\section{Related Work}
\noindent \textbf{Hateful Meme Classification \xspace} Hateful meme classification has gained prominence as an emerging multimodal task, especially thanks to Facebook's organization of the Hateful Memes Challenge (HMC) \cite{kiela2020hateful}, which established a benchmark dataset. HMC provided some baselines as a comparison for the competitors, such as VilBERT \cite{lu2019vilbert} and VisualBERT \cite{li2019visualbert}. The challenge report \cite{kiela2021hateful} showed how all the top five submissions \cite{sandulescu2020detecting, zhu2020enhance, muennighoff2020vilio, velioglu2020detecting, lippe2020multimodal} outperformed the baselines, but mainly thanks to the use of external data, additional input features and/or ensemble models.

After the end of the competition, other methods have been presented. For instance, \cite{cao2023prompting} proposes PromptHate, a prompt-based model that prompts pre-trained language models to perform hateful meme classification. Most similar to our work is Hate-CLIPper \cite{kumar2022hate}, which explicitly models the cross-modal interactions between the CLIP image and text features through intermediate fusion, thanks to a feature interaction matrix. Similarly to Hate-CLIPper, we rely on CLIP, but we introduce a 2-stage training approach that involves a Combiner network to fuse the features and the use of textual inversion to obtain a multimodal representation within the textual embedding space.

\noindent \textbf{Textual Inversion \xspace} Textual inversion has emerged as a powerful approach for the personalized image generation task in the domain of text-to-image synthesis \cite{gal2022image, kumari2023multi, ruiz2023dreambooth}. For instance, \cite{gal2022image} employs the reconstruction loss of a latent diffusion model to carry out textual inversion.

More recently, textual inversion has also been employed in virtual try-on \cite{morelli2023ladi} and personalized \cite{cohen2022my} and composed image retrieval \cite{saito2023pic2word, baldrati2023zero}. In particular, \cite{baldrati2023zero} proposes SEARLE, an approach that involves training a textual inversion network $\phi$ with a distillation-based loss. 

\section{Proposed Approach}

\begin{figure*}
    \centering
    \includegraphics[width=0.9\textwidth]{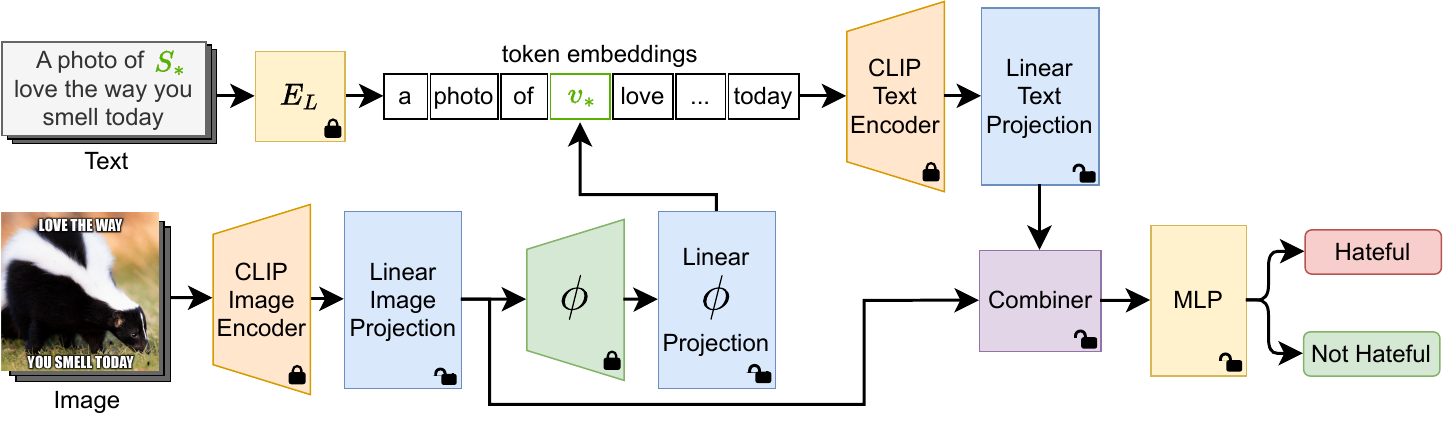}
    \vspace{-0.2ex}
    \caption{Overview of the proposed approach. We disentangle CLIP common embedding space via linear projections. We employ textual inversion to make the textual representation multimodal. We fuse the textual and visual features with a Combiner architecture. $E_{L}$ represents the CLIP embedding lookup layer. $\phi$ indicates the SEARLE textual inversion network \cite{baldrati2023zero}.}  \vspace{-1ex}
    \label{fig:architecture}
\end{figure*}

\subsection{Preliminaries}
\noindent \textbf{CLIP \xspace} The CLIP \cite{radford2021learning} vision-language model is designed to align visual and textual data within a common embedding space. It comprises a visual encoder denoted as $V_{E}$ and a text encoder denoted as $T_{E}$. These encoders extracts feature representations $V_{E}(I) \in \mathbb{R}^{d}$ and $T_{E}(E_L(Y)) \in \mathbb{R}^{d}$ from an input image $I$ and its corresponding text caption $Y$, respectively. Here, $d$ represents the dimension of the CLIP embedding space, and $E_L$ signifies the embedding lookup layer that maps each tokenized word in $Y$ to the CLIP token embedding space $\mathcal{W}$. 

\noindent \textbf{SEARLE \xspace}
SEARLE~\cite{baldrati2023zero} is an approach that involves training a textual inversion network, denoted as $\phi$, by distilling knowledge from an optimization-based method. This network exhibits the remarkable capability of efficiently performing textual inversion in a single forward pass.
Given an image $I$, the $\phi$ network maps its CLIP visual features, represented by $ V_{E}(I) \in \mathbb{R}^{d}$, into a pseudo-word token $ v_* $ within the CLIP token embedding space $\mathcal{W}$, such that $ v_* = \phi(V_{E}(I))\in \mathcal{W} $. The pre-training of the textual inversion network $\phi$ aims to ensure that the pseudo-word token $ v_* $ not only captures the visual information of $I$ but also enables effective interactions with actual words.

\subsection{\method}

\Cref{fig:architecture} shows an overview of the approach. \method focuses on three main areas: (1) enhancing the textual representation of the meme through textual inversion; (2) adapting the embedding spaces of the pre-trained model using linear projections; (3) employing an expressive multimodal fusion function.

\noindent \textbf{Enhancing textual representation \xspace}
As reported in \cite{radford2021learning}, the CLIP model exhibits strong performance in vision-language semantic tasks by utilizing only an image overlaid with some text. This underscores the efficacy of the CLIP visual encoder capabilities in directly extracting meaningful textual semantic information from raw pixel data, thus creating a powerful multimodal representation. 
To further enhance the expressiveness of the approach, \method aims to integrate the visual information of a meme within the textual embedding space, thereby generating a multimodal representation also within this domain.

To this end, we employ the textual inversion technique~\cite{gal2022image}, which involves mapping the image of a meme $I_M$ to a pseudo-word token $v_*$ residing in the CLIP token embedding space $\mathcal{W}$. To achieve this mapping efficiently and effectively, we leverage the pre-trained textual inversion network $\phi$ proposed in SEARLE~\cite{baldrati2023zero}.
Let $S_*$ represent the pseudo-word associated with $v_*$. Instead of relying solely on the text of the meme \{\textit{meme text}\}, we compute the textual features from the following prompt: ``a photo of $S_*$, \{\textit{meme text}\}".
Remarkably, thanks to the textual inversion process, these text features encompass both textual and visual information, offering a powerful multimodal representation of the meme.

\noindent \textbf{Adapting the embedding spaces \xspace} The pre-training tasks of both $\phi$ and CLIP aim to align text and image representations in a shared multimodal latent space. However, this alignment is not suitable for our task, as memes often contain text and images with different meanings. Thus, it is crucial to disentangle image and text representations in the shared multimodal latent space.
To achieve this, following \cite{kumar2022hate} we train linear projections after freezing the $\phi$ and CLIP encoders to adapt the embedding spaces. This way, we effectively disentangle the image and text modalities, enabling the learning of two distinct embedding spaces that are better suited for our specific task.

To effectively adapt the embedding spaces to our task, we follow \cite{baldrati2022conditioned} by adopting a two-stage training strategy. In the first stage, we pre-train the linear projection of the CLIP visual encoder using the same approach and architecture described in \cite{kumar2022hate}. This step allows us to achieve the desired adaptation while retaining the valuable prior knowledge captured in the pre-trained encoders.
However, pre-training the textual encoder projection layer using the same approach is not optimal due to the presence of textual inversion, since the pseudo-word tokens generated by $\phi$ would not be incorporated in the pre-training phase. 

To address this issue we introduce a second stage of training, where we train the textual and $\phi$ projection layers in conjunction with the multimodal fusion function.

\noindent \textbf{Multimodal fusion function \xspace}
Inspired by \cite{baldrati2022effective}, we adopt the Combiner network as the multimodal feature fusion function in \method. Combiner is designed to take two multimodal representations as input. The first representation originates from the pre-trained projection of the CLIP visual encoder. The second representation stems from the projection of the output of the CLIP text encoder augmented with the textual inversion, resulting in a multimodal embedding within the projected CLIP textual embedding space.
We argue that utilizing two multimodal representations empowers the Combiner and enhances its capability to capture the nuanced semantics of memes. The primary objective of the Combiner is to produce a meaningful combined representation, which then serves as input to an MLP for performing the final classification.

\section{Experimental Results}
For all the experiments, we employ the ViT-L/14 as the backbone for CLIP. The whole system is trained with a standard binary cross-entropy loss.

\subsection{Datasets}
We employ two datasets for the experiments: HMC \cite{kiela2020hateful} and HarMeme \cite{pramanick2021detecting}. HMC was proposed by Facebook for the Hateful Memes Challenge and contains 8500, 500, and 2000 synthetic memes in the training, development seen and test unseen sets, respectively. The hatred in HMC is aimed mainly at religion, race, disability, and sex. HarMeme is composed of COVID-19-related real-world memes labeled with three classes: \textit{very harmful}, \textit{partially harmful}, and \textit{harmless}. Similarly to \cite{cao2023prompting}, we combine the first two classes in a single \textit{hateful} class. HarMeme contains 3013 and 354 memes for the training and test sets, respectively.

\subsection{Quantitative Results}
\begin{table}[t!]
    \centering
    \Large
    \resizebox{0.75\linewidth}{!}{ 
    \begin{tabular}{lcc}
    \toprule
    Method & Acc. & AUROC \\
    \midrule
    CLIP Text-Only$^{\dagger}$ & 63.50 & 63.43 \\
    CLIP Image-Only$^{\dagger}$ & 74.65 & 81.35 \\
    CLIP Text + textual inversion$^{\dagger}$ & 76.65 & 83.31 \\
    CLIP Sum$^{\dagger}$ & 76.80 & 82.92 \\
    VisualBERT COCO \cite{li2019visualbert} & 69.95 & 74.59  \\
    ViLBERT CC \cite{lu2019vilbert} & 70.03 & 72.78 \\
    HMC $2^{nd}$ prize \cite{muennighoff2020vilio} & 69.50 & 83.10 \\
    HMC $1^{st}$ prize \cite{zhu2020enhance} & 73.20 & 84.49 \\
    Hate-CLIPper$^{\dagger}$ \cite{kumar2022hate} & 77.15 & 84.36 \\
    \rowcolor{tabhighlight}
    \textbf{\method} & \textbf{77.70} & \textbf{85.51} \\
    \bottomrule
    \end{tabular}}
    \vspace{0.5ex}
    \caption{Quantitative results for the HMC test unseen set. Best scores are highlighted in bold. Methods marked with $^{\dagger}$ were evaluated by us since the corresponding results are not available.}
    \label{tab:results_hmc}
\end{table}

\begin{table}[t!]
    \centering
    \Large
    \resizebox{0.65\linewidth}{!}{ 
    \begin{tabular}{lcc}
    \toprule
    Method & Acc. & AUROC \\
    \midrule
    CLIP Text-Only$^{\dagger}$ & 79.38 & 87.00 \\
    CLIP Image-Only$^{\dagger}$ & 83.90 & 91.59 \\
    CLIP Sum$^{\dagger}$ & 81.92 & 91.62\\
    DisMultiHate \cite{lee2021disentangling} & 81.24 & 86.39 \\
    PromptHate \cite{cao2023prompting} & \textbf{84.47} & 90.96 \\
    Hate-CLIPper$^{\dagger}$ \cite{kumar2022hate} & 83.90 & 91.87 \\
    \rowcolor{tabhighlight}
    \textbf{\method} & 81.64 & \textbf{92.83} \\
    \bottomrule
    \end{tabular}}
    \vspace{0.5ex}
    \caption{Quantitative results for the HarMeme test set. Best scores are highlighted in bold. Methods marked with $^{\dagger}$ were evaluated by us since the corresponding results are not available.}
    \label{tab:results_harmeme}
\end{table}

We compare \method with baselines and competing methods in \cref{tab:results_hmc,tab:results_harmeme} for the HMC test unseen and HarMeme test sets, respectively. We do not consider DisMultiHate \cite{lee2021disentangling} and PromptHate \cite{cao2023prompting} for the HMC dataset because the original papers report only the results on the dev seen split and the pre-trained models are not public. We evaluate the performance of the models using the AUROC \cite{bradley1997use} -- which was the primary evaluation metric for Facebook's Hateful Memes Challenge -- and the accuracy.

The presence of confounders renders the HMC dataset particularly challenging for unimodal methods and necessitates robust multimodal reasoning, as evidenced by the poor results obtained by CLIP Text-Only. 
However, employing only the CLIP image encoder yields significantly better results, demonstrating its capability to extract representations that capture the semantic meaning of both the image and the overlaid text.
When we enhance the CLIP textual encoder using the textual inversion (see CLIP Text + textual inversion in \cref{tab:results_hmc}), we successfully extract semantic visual features from memes and transfer them into CLIP textual embedding space. This process yields a meaningful multimodal representation, improving the results obtained by the visual encoder and the sum of visual and textual features.

\method outperforms all the baselines and prize winners of the challenge. 
When compared to the current SotA approach Hate-CLIPper~\cite{kumar2022hate}, the proposed method shows a significant improvement. Since both models employ the same CLIP backbone, the increase in the metrics is attributable to the effectiveness and impact of the novel components that constitute our approach.

\Cref{tab:results_harmeme} shows the results for the HarMeme dataset.
Differently from the HMC dataset, utilizing a uni-modal architecture that considers only the text of the meme obtains good results. Consequently, this dataset does not challenge multimodal reasoning to the same extent as the HMC one.  However, since the memes are directly collected "in the wild" from social networks, they may exhibit lower quality, but they provide valuable insights into real-world behaviors and interactions.
Therefore, the experiments on this dataset offer an excellent opportunity to assess the generalization capabilities of \method.
Remarkably, even in this scenario, our model outperforms all existing SotA architectures.

\subsection{Ablation Study} 
We present an ablation study on the HMC dataset in \cref{tab:ablation}. We start with the Hate-CLIPper architecture \cite{kumar2022hate} as the base model and gradually integrate the main parts of \method: the Combiner network, 2-stage training, and textual inversion. Finally, we measure their relative importance.

Replacing the fusion method in Hate-CLIPper with the Combiner network results in a performance increase. The Combiner enhances the expressiveness of the fusion function, thus improving its ability to model the semantic interaction between text and image representations. The improvement observed with the two-stage training process underscores the importance of carefully adapting the embedding spaces to our specific task by disentangling the image and text representation. Finally, by using textual inversion, we provide two meaningful multimodal representations to the Combiner network in both the visual and textual embedding spaces, which enables better modeling of the semantic meaning of the meme.

\begin{table}[!t]
    \centering
    \Large
    \resizebox{0.8\linewidth}{!}{ 
    \begin{tabular}{l ccc cc}
    \toprule
    Method & Combiner & 2-stage & TI & Acc. & AUROC \\ \midrule
    Base &  &  &  & 77.15 & 84.36 \\
         & \cmark &  &  & \textbf{77.80} & 84.68 \\
         & \cmark & \cmark &  & 77.55 & 85.05 \\
    \rowcolor{tabhighlight}
    \textbf{\method} & \cmark & \cmark & \cmark & 77.70 & \textbf{85.51} \\
    \bottomrule
    \end{tabular}}
    \vspace{0.5ex}
    \caption{Ablation study on the test unseen set of the HMC dataset. TI stands for Textual Inversion.}
    \label{tab:ablation}
\end{table}

\section{Conclusion}
In this paper, we introduce \method, a novel method for classifying hateful memes. Our approach leverages CLIP and textual inversion to generate meaningful multimodal representations in both textual and visual domains. To adapt the pre-trained models, we integrate trainable linear projections into our architecture and adopt a two-stage training strategy. Also, we enhance the expressiveness of the fusion function through a Combiner network, thus improving the modeling of semantic interactions between the meme representations. Experiments on the HMC and HarMeme datasets show that \method achieves SotA results.

\vspace{-2ex}
\paragraph{Acknowledgments}
This work was partially supported by the European Commission under European Horizon 2020 Programme, grant number 101004545 - ReInHerit.

{\small
\bibliographystyle{ieee_fullname}
\bibliography{egbib}

\begin{thebibliography}{10}\itemsep=-1pt

\bibitem{baldrati2023zero}
Alberto Baldrati, Lorenzo Agnolucci, Marco Bertini, and Alberto Del~Bimbo.
\newblock Zero-shot composed image retrieval with textual inversion.
\newblock {\em arXiv preprint arXiv:2303.15247}, 2023.

\bibitem{baldrati2022conditioned}
Alberto Baldrati, Marco Bertini, Tiberio Uricchio, and Alberto Del~Bimbo.
\newblock Conditioned and composed image retrieval combining and partially fine-tuning clip-based features.
\newblock In {\em Proceedings of the IEEE/CVF Conference on Computer Vision and Pattern Recognition}, pages 4959--4968, 2022.

\bibitem{baldrati2022effective}
Alberto Baldrati, Marco Bertini, Tiberio Uricchio, and Alberto Del~Bimbo.
\newblock Effective conditioned and composed image retrieval combining clip-based features.
\newblock In {\em Proceedings of the IEEE/CVF Conference on Computer Vision and Pattern Recognition}, pages 21466--21474, 2022.

\bibitem{bradley1997use}
Andrew~P Bradley.
\newblock The use of the area under the roc curve in the evaluation of machine learning algorithms.
\newblock {\em Pattern recognition}, 30(7):1145--1159, 1997.

\bibitem{cao2023prompting}
Rui Cao, Roy Ka-Wei Lee, Wen-Haw Chong, and Jing Jiang.
\newblock Prompting for multimodal hateful meme classification.
\newblock {\em arXiv preprint arXiv:2302.04156}, 2023.

\bibitem{cohen2022my}
Niv Cohen, Rinon Gal, Eli~A Meirom, Gal Chechik, and Yuval Atzmon.
\newblock “this is my unicorn, fluffy”: Personalizing frozen vision-language representations.
\newblock In {\em European Conference on Computer Vision}, pages 558--577. Springer, 2022.

\bibitem{gal2022image}
Rinon Gal, Yuval Alaluf, Yuval Atzmon, Or Patashnik, Amit~H Bermano, Gal Chechik, and Daniel Cohen-Or.
\newblock An image is worth one word: Personalizing text-to-image generation using textual inversion.
\newblock {\em arXiv preprint arXiv:2208.01618}, 2022.

\bibitem{kiela2021hateful}
Douwe Kiela, Hamed Firooz, Aravind Mohan, Vedanuj Goswami, Amanpreet Singh, Casey~A Fitzpatrick, Peter Bull, Greg Lipstein, Tony Nelli, Ron Zhu, et~al.
\newblock The hateful memes challenge: Competition report.
\newblock In {\em NeurIPS 2020 Competition and Demonstration Track}, pages 344--360. PMLR, 2021.

\bibitem{kiela2020hateful}
Douwe Kiela, Hamed Firooz, Aravind Mohan, Vedanuj Goswami, Amanpreet Singh, Pratik Ringshia, and Davide Testuggine.
\newblock The hateful memes challenge: Detecting hate speech in multimodal memes.
\newblock {\em Advances in Neural Information Processing Systems}, 33:2611--2624, 2020.

\bibitem{kumar2022hate}
Gokul~Karthik Kumar and Karthik Nanadakumar.
\newblock Hate-clipper: Multimodal hateful meme classification based on cross-modal interaction of clip features.
\newblock {\em arXiv preprint arXiv:2210.05916}, 2022.

\bibitem{kumari2023multi}
Nupur Kumari, Bingliang Zhang, Richard Zhang, Eli Shechtman, and Jun-Yan Zhu.
\newblock Multi-concept customization of text-to-image diffusion.
\newblock In {\em Proceedings of the IEEE/CVF Conference on Computer Vision and Pattern Recognition}, pages 1931--1941, 2023.

\bibitem{lee2021disentangling}
Roy Ka-Wei Lee, Rui Cao, Ziqing Fan, Jing Jiang, and Wen-Haw Chong.
\newblock Disentangling hate in online memes.
\newblock In {\em Proceedings of the 29th ACM International Conference on Multimedia}, pages 5138--5147, 2021.

\bibitem{li2019visualbert}
Liunian~Harold Li, Mark Yatskar, Da Yin, Cho-Jui Hsieh, and Kai-Wei Chang.
\newblock Visualbert: A simple and performant baseline for vision and language.
\newblock {\em arXiv preprint arXiv:1908.03557}, 2019.

\bibitem{lippe2020multimodal}
Phillip Lippe, Nithin Holla, Shantanu Chandra, Santhosh Rajamanickam, Georgios Antoniou, Ekaterina Shutova, and Helen Yannakoudakis.
\newblock A multimodal framework for the detection of hateful memes.
\newblock {\em arXiv preprint arXiv:2012.12871}, 2020.

\bibitem{lu2019vilbert}
Jiasen Lu, Dhruv Batra, Devi Parikh, and Stefan Lee.
\newblock Vilbert: Pretraining task-agnostic visiolinguistic representations for vision-and-language tasks.
\newblock {\em Advances in neural information processing systems}, 32, 2019.

\bibitem{morelli2023ladi}
Davide Morelli, Alberto Baldrati, Giuseppe Cartella, Marcella Cornia, Marco Bertini, and Rita Cucchiara.
\newblock Ladi-vton: Latent diffusion textual-inversion enhanced virtual try-on.
\newblock {\em arXiv preprint arXiv:2305.13501}, 2023.

\bibitem{muennighoff2020vilio}
Niklas Muennighoff.
\newblock Vilio: State-of-the-art visio-linguistic models applied to hateful memes.
\newblock {\em arXiv preprint arXiv:2012.07788}, 2020.

\bibitem{pramanick2021detecting}
Shraman Pramanick, Dimitar Dimitrov, Rituparna Mukherjee, Shivam Sharma, Md Akhtar, Preslav Nakov, Tanmoy Chakraborty, et~al.
\newblock Detecting harmful memes and their targets.
\newblock {\em arXiv preprint arXiv:2110.00413}, 2021.

\bibitem{radford2021learning}
Alec Radford, Jong~Wook Kim, Chris Hallacy, Aditya Ramesh, Gabriel Goh, Sandhini Agarwal, Girish Sastry, Amanda Askell, Pamela Mishkin, Jack Clark, et~al.
\newblock Learning transferable visual models from natural language supervision.
\newblock In {\em International conference on machine learning}, pages 8748--8763. PMLR, 2021.

\bibitem{ruiz2023dreambooth}
Nataniel Ruiz, Yuanzhen Li, Varun Jampani, Yael Pritch, Michael Rubinstein, and Kfir Aberman.
\newblock Dreambooth: Fine tuning text-to-image diffusion models for subject-driven generation.
\newblock In {\em Proceedings of the IEEE/CVF Conference on Computer Vision and Pattern Recognition}, pages 22500--22510, 2023.

\bibitem{saito2023pic2word}
Kuniaki Saito, Kihyuk Sohn, Xiang Zhang, Chun-Liang Li, Chen-Yu Lee, Kate Saenko, and Tomas Pfister.
\newblock Pic2word: Mapping pictures to words for zero-shot composed image retrieval.
\newblock In {\em Proceedings of the IEEE/CVF Conference on Computer Vision and Pattern Recognition}, pages 19305--19314, 2023.

\bibitem{sandulescu2020detecting}
Vlad Sandulescu.
\newblock Detecting hateful memes using a multimodal deep ensemble.
\newblock {\em arXiv preprint arXiv:2012.13235}, 2020.

\bibitem{velioglu2020detecting}
Riza Velioglu and Jewgeni Rose.
\newblock Detecting hate speech in memes using multimodal deep learning approaches: Prize-winning solution to hateful memes challenge.
\newblock {\em arXiv preprint arXiv:2012.12975}, 2020.

\bibitem{zhu2020enhance}
Ron Zhu.
\newblock Enhance multimodal transformer with external label and in-domain pretrain: Hateful meme challenge winning solution.
\newblock {\em arXiv preprint arXiv:2012.08290}, 2020.

\end{thebibliography}
}

\end{document}